\documentclass{article}

\usepackage[preprint]{neurips_2026}

\usepackage[utf8]{inputenc} 
\usepackage[T1]{fontenc}    
\usepackage{hyperref}       
\usepackage{url}            
\usepackage{booktabs}       
\usepackage{amsfonts}       
\usepackage{nicefrac}       
\usepackage{microtype}      
\usepackage{xcolor}         
\usepackage{graphicx}

\usepackage{amsmath,amssymb}
\usepackage{amsfonts}
\usepackage{nicefrac}
\usepackage{microtype}
\usepackage{xcolor}
\usepackage{graphicx}
\usepackage{wrapfig}
\usepackage{multirow}
\usepackage{subcaption}
\usepackage{wrapfig}
\usepackage{listings}
\newcommand{\sysname}{AgentServeSim}
\title{AgentServeSim: Serving-System Simulation and Policy Search for LLM Agent Programs}

\author{%
  Rakibul Hasan Rajib \quad Mengxin Zheng \quad Qian Lou \\
  University of Central Florida \\
  \texttt{\{rakibul,  mengxin.zheng, qian.lou\}@ucf.edu}
}

\begin{document}

\maketitle

\begin{abstract}
Large language model agents execute programs comprising multiple model turns interleaved with external tool calls. Their job completion time depends on how the serving system retains KV state across tool gaps, routes successor turns, and schedules competing programs. Most existing serving simulators operate on request streams in which arrivals are externally supplied and KV state follows request- or cache-scoped semantics. They therefore cannot jointly represent the cross-turn state and policy-dependent successor releases needed to evaluate counterfactual agent-serving trajectories. We present \sysname{}, a simulator whose unit of execution is the agent program. A Program Control Block maintains cross-turn state, while a Program Orchestrator causally releases successor turns from simulated predecessor completions. A Retention Plane controls KV state across tool gaps, and a Dispatch Plane determines where and when each ready turn executes. We validate \sysname{} against real vLLM deployments in 20 paired simulator--real cells spanning two GPU platforms, Llama-3.1-8B and 70B, coding and function-calling agents, and five arrival rates. Mean JCT error remains within 5.5\% on B200 and 5.2\% in the saturated RTX PRO 6000 regime. Finally, we propose LLM-driven automated agent-serving policy search using \sysname{} as a CPU-based fitness evaluator. The resulting policies improve mean JCT over hand-written seed policies by 0.5\% for KV retention and 2.8\% for scheduling.
\end{abstract}
\section{Introduction}
\label{sec:intro}

Large language models (LLMs) are increasingly deployed as agents that accomplish tasks through multiple model calls interleaved with external tool executions~\citep{yao2023react,jimenez2024swebench,patil2025bfcl}. These agents power coding assistants, web agents, and long-horizon planners, but serving them efficiently is expensive. Their user-facing performance is determined by the job completion time (JCT) of the entire agent program rather than the latency of any single model request. Improving JCT requires deciding what happens to a program's KV cache while it waits for a tool, where its next turn executes, and when that turn is scheduled.

At the core of an agent workload is a program that alternates between LLM turns and tool calls. A program arrives, executes an LLM turn, waits for the requested tool, and releases its successor turn only after both complete. Three properties make this lifecycle different from a stream of independent requests. First, tool calls introduce gaps with a heavy-tailed duration distribution: in production traces of two coding agents, calls longer than one minute constitute only 4\% of tool calls but account for 85\% of total tool time~\citep{tracelab2026}. Second, successive turns commonly extend a shared context, creating substantial opportunities for cross-turn KV reuse; the same traces report a global prefix-cache hit rate of 95.7\%~\citep{tracelab2026}. Third, these effects accumulate over many turns: programs in our SWE-Bench traces contain a median of 47 turns (\S\ref{app:workloads}). Consequently, a decision made at one turn can affect memory pressure, cache reuse, and completion times throughout the remainder of the program.

Serving systems have introduced policies specialized for these decisions, including KV retention across tool gaps~\citep{abhyankar2024infercept,li2026continuum}, session-aware routing~\citep{smetric2026}, and program-aware scheduling~\citep{luo2025autellix}. However, evaluating such policies by repeatedly modifying and deploying a real serving engine is prohibitively expensive. Sweeping deployment configurations for a single model has been estimated to require 42K GPU-hours~\citep{agrawal2024vidur}, while agent serving adds interacting retention, routing, and scheduling dimensions. Some configurations of interest, such as memory tiers with tunable capacity and bandwidth, may not be physically available. Simulation is therefore a natural instrument for exploring this design space and, by reducing each evaluation from GPU-hours to CPU minutes, for searching serving policies automatically.

In this paper, we observe that existing serving simulators~\citep{agrawal2024vidur,patke2024apex,cho2024llmservingsim,cho2026llmservingsim2,feng2025frontier} lack a general program-level abstraction for agent workloads. Most treat each request as the complete unit of arrival, execution state, and KV lifetime. Frontier~\cite{feng2025frontier} introduces causally released stateful request chains, but fixes each successor to its predecessor's replica and relies on the engine's native prefix-cache behavior~\citep{feng2025frontier}. Agent programs instead maintain state and causal dependencies across requests. This mismatch limits existing simulators in two ways.

First, existing simulators fix the disposition of KV state after a request completes. They may release it immediately, retain it as anonymous content-keyed cache entries under a fixed replacement policy~\citep{cho2026llmservingsim2}, or leave it in an engine-managed prefix cache on the same replica~\citep{feng2025frontier}. In an agent-serving system, however, the tool gap is itself a policy decision point. The program's KV state may remain in device memory, migrate to another memory tier, or be evicted and later recomputed. The appropriate decision depends on the expected gap, memory pressure, recomputation cost, and available transfer bandwidth~\citep{abhyankar2024infercept,li2026continuum}. Evaluating these choices requires program ownership of KV state and policy-controlled retention across requests.

Second, existing simulators generally treat each request's arrival as an external input and make routing and scheduling decisions from request-scoped state. Agent programs instead couple these decisions across time. Programs arrive exogenously, but a successor turn becomes ready only after its predecessor completes under the evaluated policy and its tool returns. Its placement determines whether its prefix is locally cached and under what memory pressure it executes, while its scheduling delay determines when the following turn can be released. A recorded trace specifies the logical program, including its turns, token counts, and tool durations, but records only one serving trajectory. Replaying its turns at recorded timestamps fixes the release schedule and cache state that a different policy is supposed to change; the counterfactual serving trajectory must instead be generated causally.

To address these limitations, we present \sysname{}, whose unit of simulation is the agent program. Inspired by an operating system process, it represents an LLM turn as a compute burst, a tool call as a blocking system call, and the program's KV cache as a working set that persists across turns. A Program Control Block maintains program identity, KV ownership, and execution history, while a Program Orchestrator releases successor turns from simulated predecessor completions. On this abstraction, the \emph{Retention Plane} determines how KV state is managed across tool gaps, and the \emph{Dispatch Plane} determines where and when each ready turn executes. Together, they generate the policy-dependent release, placement, and cache-residency trajectory of a fixed logical program.

We validate \sysname{} against real vLLM deployments in 20 paired simulator--real cells spanning two GPU platforms, Llama-3.1-8B and 70B, coding and function-calling agents, and five arrival rates. Mean JCT error remains within 5.5\% on B200 and 5.2\% in the saturated RTX PRO 6000 regime. An LLM-driven policy search further improves mean JCT over hand-written seed policies by 0.5\% for KV retention and 2.8\% for scheduling.

In summary, this paper makes the following contributions:

\begin{itemize}
\item We introduce a program-level simulation abstraction comprising a Program Control Block and Retention and Dispatch Planes, under which successor arrivals and cached state are outcomes of simulated execution rather than fixed trace inputs.

\item We validate \sysname{} against real vLLM deployments across GPU platforms, model scales, agent workloads, arrival rates, and memory-pressure regimes.

\item We propose LLM-driven automated agent-serving policy search using \sysname{} as a CPU-based fitness evaluator, discovering KV-retention and scheduling policies that outperform hand-written seed policies.
\end{itemize}

\section{Related Work}
\label{sec:related}

\textbf{Serving simulators.}
Vidur~\citep{agrawal2024vidur}, APEX~\citep{patke2024apex}, and LLMServingSim~\citep{cho2024llmservingsim,cho2026llmservingsim2}, which \sysname{} extends, model request streams using profiled operator costs and validate against engines serving independent requests. Frontier~\citep{feng2025frontier} is closest: it models stateful request chains and releases successors causally, capturing time coupling, but pins each successor to its predecessor's replica and leaves retention to the engine's native prefix cache. Revati~\citep{agrawal2026revati} reduces simulator--engine policy drift through engine emulation; our parity checker instead compares the decisions made by the same program-level policy in simulation and real execution. These simulators do not jointly expose and validate program-level retention, routing, and scheduling under sustained KV-pool pressure.

\textbf{Agent-serving systems.}
InferCept~\citep{abhyankar2024infercept} manages KV state across tool gaps; Continuum~\citep{li2026continuum} co-designs KV continuity with program-FCFS scheduling; Autellix~\citep{luo2025autellix} orders requests by attained program service; and SMetric~\citep{smetric2026} routes requests using cached-prefix residency and load. These systems demonstrate the value of program-level policies but evaluate particular designs in production engines, making cross-policy counterfactuals expensive to isolate. \sysname{} separates retention, routing, and scheduling behind a common program-state abstraction, allowing their components to be composed and compared in simulation (Section~\ref{sec:usecase}).

\section{\sysname{} Design}
\label{sec:design}

\begin{figure}[t]
\centering
\includegraphics[width=\columnwidth]{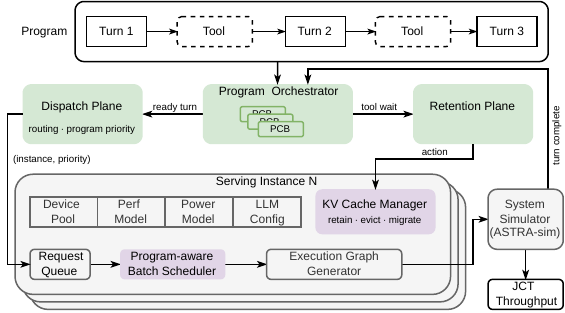}
\caption{\sysname{} architecture. The Program Orchestrator and Program Control Blocks maintain execution across calls; the Dispatch Plane controls routing and scheduling; and the Retention Plane controls KV state across tool gaps.}
\label{fig:arch}
\end{figure}

\subsection{System Overview}

Figure~\ref{fig:arch} shows \sysname{} from the program's perspective. The workload specifies each program as an ordered sequence of LLM calls with their token counts and the tool durations between them; it fixes the logical program but not its serving trajectory. When a program arrives, the Program Orchestrator creates a Program Control Block (PCB) that maintains all of its state across calls. When a call is ready, the Orchestrator sends it to the Dispatch Plane, which selects a serving instance and assigns a scheduling priority. Each instance contains a device pool, a performance model, a power model, and a KV cache manager. Its batch scheduler forms a batch from the queued calls according to their priorities, the KV cache manager resolves prefix reuse against the resident state, and the execution graph generator emits an execution graph for the batch. The System Simulator evaluates the graph, accounting for computation, communication, synchronization, and memory access. When the call completes, the Retention Plane assigns its KV state a value and a protection deadline, which the KV cache manager honors when reclaiming memory under pressure, and the Orchestrator holds the program for its tool duration before releasing the next call. The loop repeats until every program has completed its last call. During execution, \sysname{} reports system-level metrics such as memory usage, energy consumption, and throughput; after completion, it reports JCT and preemptions per program and TTFT, TPOT, queueing delay, and end-to-end latency per call.

All policies operate on the same observable state. Let $x_p(t)$ denote the state of program $p$ and $s(t)$ the deployable system signals at time $t$, including per-instance load and KV-pool utilization. The two policy planes expose the following interface:
\begin{equation}
\begin{aligned}
(i_{p,k},q_{p,k}) &= \pi_{\mathrm{dispatch}}\!\left(x_p(r_{p,k}),s(r_{p,k})\right),\\
(v_{p,k},d_{p,k}) &= \pi_{\mathrm{retain}}\!\left(x_p(c_{p,k}),s(c_{p,k})\right),
\end{aligned}
\label{eq:interface}
\end{equation}
where $r_{p,k}$ and $c_{p,k}$ are the release and completion times of call $k$, $i_{p,k}$ is its assigned instance, $q_{p,k}$ its scheduling priority, and $v_{p,k}$ and $d_{p,k}$ the value and protection deadline of its KV state. Policies return only these declarative decisions; batching, preemption, memory reclamation, and execution timing remain engine mechanisms.

\subsection{Program Orchestrator}

A request-level simulator reads every request arrival from the workload and discards its execution state at completion. This model cannot represent an agent program: a successor becomes ready only after its predecessor completes and its tool returns, while the state required by program-aware policies persists across calls. Replaying recorded per-call timestamps would fix the timeline that the evaluated policy is supposed to change.

The Program Orchestrator addresses both limitations by maintaining one Program Control Block (PCB) for every live program. Let $a_p$ be the arrival time of program $p$ and $g_{p,k}$ the tool duration following call $k$. The Orchestrator releases calls according to
\begin{equation}
r_{p,1}=a_p,
\qquad
r_{p,k+1}=c_{p,k}+g_{p,k}.
\label{eq:causal-release}
\end{equation}
Because $c_{p,k}$ includes policy-dependent queueing and execution, Equation~\ref{eq:causal-release} generates the successor-release schedule induced by the policy being evaluated.

The PCB stores program identity and position, attained service, prior placement, KV residency and protection state, and current tool state. It is updated at call release, call completion, and memory-pressure events. The PCB also defines a causal observation boundary: future output lengths and tool durations stored in the trace are excluded from $x_p(t)$ and $s(t)$ until revealed by execution. The Orchestrator therefore provides the persistent, non-anticipatory program state required by both policy planes.

\subsection{Dispatch Plane}

Request-scoped dispatch cannot use a program's prior placement or cumulative service. Consequently, it cannot express routing based on prior program placement or scheduling based on cumulative program service. These decisions also affect future execution: routing determines cache locality and instance pressure, while queueing determines when the successor call can be released.

The Dispatch Plane jointly exposes routing and scheduling through the first line of Equation~\ref{eq:interface}. Routing is evaluated first because the selected instance determines both the destination queue and whether the program's prefix is resident. We instantiate round-robin, least-loaded, and session-affinity routing; session affinity reuses the program's prior instance unless a capacity threshold triggers fallback to the least-loaded instance.

Within an instance, calls with smaller $q_{p,k}$ execute first, with release time breaking ties. Let
\begin{equation}
A_p(t) = \sum_{j:\,c_{p,j}\leq t} C_{p,j}
\label{eq:attained-service}
\end{equation}
denote the service attained by program $p$, where $C_{p,j}$ is the wall-clock time of call $j$ on its instance from first admission to completion, excluding the initial queueing delay but including any preemption interval, counted once per instance regardless of its tensor-parallel degree. The evaluated scheduling policies instantiate
\begin{equation}
q_{p,k} =
\begin{cases}
r_{p,k}, & \text{FCFS},\\
a_p, & \text{Program-FCFS},\\
A_p(r_{p,k}), & \text{PLAS}.
\end{cases}
\label{eq:scheduling}
\end{equation}
Program-FCFS preserves program arrival order, following Continuum~\citep{li2026continuum}; PLAS prioritizes programs that have consumed less service, following Autellix~\citep{luo2025autellix}. The Dispatch Plane thus makes placement and ordering functions of persistent program state while leaving execution mechanisms unchanged.

\subsection{Retention Plane}

Request-level simulators typically release KV state at request completion or retain it under a fixed cache-replacement rule. Neither choice can represent agent-serving policies that adapt retention to the expected tool gap, memory pressure, or recomputation cost~\citep{abhyankar2024infercept,li2026continuum}. As a result, the lifetime of cross-turn KV state remains fixed rather than evaluated as a policy decision.

The Retention Plane addresses this limitation through the second line of Equation~\ref{eq:interface}. At call completion, the policy assigns the program's KV state a value $v_{p,k}$ and protection deadline $d_{p,k}$. Higher values indicate greater benefit from retention. Under memory pressure, the memory manager forms the eviction key
\begin{equation}
e_{p,k}(t) = \left(\mathbb{I}[t<d_{p,k}],\; v_{p,k}\right)
\label{eq:eviction}
\end{equation}
and reclaims entries in ascending lexicographic order until sufficient capacity is available. Expired entries are considered first, followed by lower-valued entries. Protection remains a preference rather than a lock: if all entries are protected, the least-valued protected entry becomes the victim. If $d_{p,k}=c_{p,k}$, the KV entry is released at completion; otherwise it remains resident and participates in pressure-driven reclamation according to Equation~\ref{eq:eviction}.

The evaluated deadline policies are
\begin{equation}
d_{p,k} = c_{p,k} +
\begin{cases}
0, & \text{immediate release},\\
\tau, & \text{fixed TTL},\\
Q_{\alpha}(G\mid z_{p,k}), & \text{per-tool TTL},\\
\tau\, g(u(c_{p,k})), & \text{pressure-aware TTL},
\end{cases}
\label{eq:retention-deadlines}
\end{equation}
where $Q_{\alpha}(G\mid z)$ is an empirical quantile of observed gaps for tool type $z$, $u(t)$ is KV-pool utilization, and $g$ is a non-increasing pressure function. For valuation, LRU sets $v_{p,k}$ from the entry's last-access time. Min-waste compares the memory-time cost of retaining the entry over the predicted gap with the re-prefill cost of discarding it, following InferCept~\citep{abhyankar2024infercept}; the shipped instance uses this comparison to choose between protection until the predicted gap elapses and release at completion, and leaves $v_{p,k}$ unset. The $g$ and min-waste instances used in the experiments are given in Appendix~\ref{app:policies}. Immediate release recovers request-scoped KV lifetime, while fixed deadline and valuation rules recover static retention semantics. The Retention Plane therefore turns cross-turn KV lifetime into a programmable policy outcome while leaving capacity enforcement to the memory manager.

\section{Experimental Setup}
\paragraph{Implementation.}
\label{sec:setup}
We build \sysname{} on LLMServingSim~2.0~\citep{cho2026llmservingsim2} because it already models the request-level mechanisms orthogonal to our contribution, including batching, KV memory management, distributed execution, and profile-based operator timing. This allows \sysname{} to add program-level execution while preserving the base simulator's serving and hardware models. We connect the program layer to LLMServingSim~2.0 and its ASTRA-Sim backend~\citep{won2023astrasim2} through four hooks: scheduling priority, retention decisions, victim protection, and KV residency. The same policy implementations execute in vLLM~\citep{kwon2023vllm} through a thin adapter.

\paragraph{Hardware platforms.}
We validate \sysname{} on NVIDIA B200 GPUs (180~GB HBM3e, NVLink), serving Llama-3.1-8B at TP1 and Llama-3.1-70B at TP2, and RTX PRO 6000 GPUs (96~GB GDDR7, PCIe), serving Llama-3.1-70B at TP2. Here, TP denotes the tensor-parallel degree: TP1 executes the model on one GPU, whereas TP2 shards it across two GPUs. The real serving system is vLLM~0.19~\citep{kwon2023vllm}, configured with the same agent KV-retention and scheduling policies as the simulator. Each cell uses a single serving instance, so the validation exercises retention and scheduling but not routing.

\paragraph{Simulator configuration.}
\sysname{} is parameterized with device specifications matching the real systems (Appendix~\ref{app:cells}). Operator-level latency profiles are collected for each (device, model) pair with the base simulator's profiler; collectives are modeled by ASTRA-Sim at the NCCL all-reduce bandwidth measured on each node pair; a per-step residual table is measured over a grid of batch shapes; and the KV pool size is taken from the engine's own allocation report.

\paragraph{Workloads.}
We use 50 SWE-bench programs collected with mini-SWE-agent~\citep{minisweagent2025} (median 47 calls) and 100 BFCL~v4 function-calling programs~\citep{patil2025bfcl} (median 5.5 calls), described in Appendix~\ref{app:workloads}. Each cell submits its programs under Poisson arrivals at $\{0.02, 0.04, 0.06, 0.08, 0.1\}$ jobs per second (JPS), where a job is one program. For each call we replay the captured prompt, generated-token count, and tool duration on both the real and the simulated side, so JCT differences reflect serving-system behavior rather than agent-side nondeterminism; release times, ordering, and residency are generated on both sides.
\section{Results}
\label{sec:results}

\subsection{Fidelity under program-aware serving policies}
\label{sec:fidelity}

Existing serving simulators are validated against engines serving independent requests under the engine's default policies. That evidence does not transfer to the setting this paper targets: with TTL retention and PLAS scheduling, the simulator's own decisions determine which contexts remain resident and which programs wait, so errors can compound over the tens of calls in a program. We therefore ask whether \sysname{} reproduces a real engine running the same policies, including when the KV pool saturates and both policies become consequential.

\begin{figure}[t]
\centering
\includegraphics[width=\linewidth]{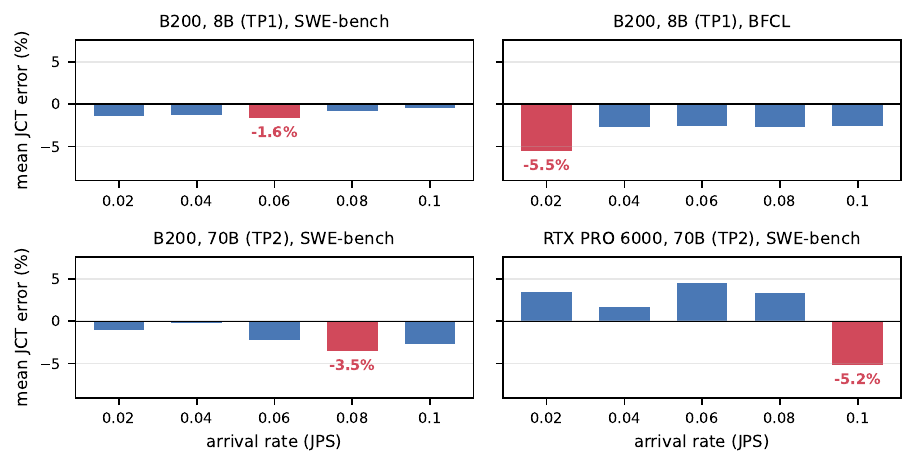}
\caption{Mean-JCT error of \sysname{} relative to real vLLM under TTL retention and PLAS scheduling, per arrival rate; the largest error in each configuration is marked. The RTX configuration is the memory-pressure regime. p99 error and per-program MAPE are in Appendix~\ref{app:cells}.}
\label{fig:validation}
\end{figure}

Figure~\ref{fig:validation} reports mean-JCT error across the 20 paired simulator--real cells. On B200, mean-JCT error stays within $\pm3.5$\% on SWE-bench. The BFCL configuration has the largest B200 error, reaching $-5.5$\% at the lowest arrival rate and falling to $-2.5$\% at the highest rate. On RTX PRO 6000, mean-JCT error ranges from $-5.2$\% to $+4.5$\%. Across these cells, p99 JCT error remains within $\pm2.7$\% on B200 and reaches $+7.1$\% on RTX PRO 6000, while per-program MAPE remains within 5.9\% and 11.2\%, respectively (Appendix~\ref{app:cells}).

Figure~\ref{fig:validation} shows the mean-JCT error across the 20 paired simulator--real cells. Both \sysname{} and vLLM use TTL retention ($\tau=2$~s) and PLAS scheduling; the complete policy configuration is reported in Appendix~\ref{app:cells}. On B200, mean-JCT error stays within $\pm3.5$\% on SWE-bench. The BFCL configuration has the largest B200 error ($-5.5$\% at the lowest rate): these programs last about 11~s, and the real engine pays a wake-up of roughly 60~ms on the first call after an idle gap, which the simulator does not model. At the highest rate, where the engine never idles, the error falls to $-2.5$\%. The RTX PRO 6000 configuration is the memory-pressure regime. Llama-3.1-70B at TP2 leaves a 114,768-token KV pool for a workload containing 38.7M context tokens, and the resulting makespan is five times the arrival span. \sysname{} reproduces this regime at all five rates: mean-JCT error ranges from $-5.2$\% to $+4.5$\%, and at 0.02 JPS the mean per-call queueing delay is 37.1~s in simulation versus 35.2~s in vLLM. Across all cells, p99 error remains within $\pm2.7$\% on B200 and reaches $+7.1$\% on RTX PRO 6000. Per-program  mean absolute percentage error (MAPE) remains within 5.9\% on B200 and 11.2\% on RTX PRO 6000; it is within 3.3\% on the B200 SWE-bench cells and ranges from 7\% to 11\% in the RTX PRO 6000 regime (Appendix~\ref{app:cells}).

\subsection{Agent-serving design-space exploration}
\label{sec:sweep}

\begin{wraptable}{r}{0.46\linewidth}
\vspace{-10pt}
\caption{Simulated sweep on RTX 70B with 50 programs at 0.02 JPS. Values are changes in mean JCT relative to to least-recently-used (LRU) retention with FCFS scheduling (3726~s); lower is better.}
\label{tab:sweep}
\centering
\scriptsize
\setlength{\tabcolsep}{4pt}
\begin{tabular}{llr}
\toprule
retention & scheduling & $\Delta$JCT \\
\midrule
TTL & FCFS & $+0.4$\% \\
pressure-aware TTL & FCFS & $+0.4$\% \\
LRU & PLAS & $-3.9$\% \\
TTL & PLAS & $-3.0$\% \\
per-tool TTL & program FCFS & $-2.8$\% \\
pressure-aware TTL & PLAS & $-5.0$\% \\
\bottomrule
\end{tabular}
\end{wraptable}

Agent serving exposes a large design space of interacting policy choices. Exploring this space on real hardware is expensive because every policy configuration requires a separate GPU run. \sysname{} provides a lower-cost, controlled alternative for systematically evaluating this design space.

Table~\ref{tab:sweep} reports a joint sweep of retention and scheduling policies at the RTX memory-pressure point. Changing retention under FCFS does not improve JCT: fixed and pressure-aware TTL both increase it by 0.4\%. Switching from FCFS to PLAS under LRU reduces JCT by 3.9\%, while combining pressure-aware TTL with PLAS reduces it by 5.0\%. Fixed TTL weakens the PLAS improvement by 0.9 percentage points, consistent with indiscriminate protection displacing KV state belonging to programs that PLAS will run next. These results show that scheduling determines which programs wait, while retention determines whether their KV state remains available when they resume. Each simulated point costs approximately one CPU-core-hour, whereas the corresponding vLLM evaluation occupies two GPUs for three to six hours.

\subsection{LLM-driven evolutionary policy search}
\label{sec:usecase}

Given the size of the agent-serving design space, designing policies by hand for every workload and hardware configuration does not scale. We therefore use \sysname{} as the evaluator in an evolutionary policy-search loop: an LLM modifies executable serving-policy code, \sysname{} evaluates the resulting policy on a complete agent workload, and the measured JCT guides subsequent modifications. Unlike parameter tuning over a predefined policy family, the search evolves the policy implementation itself while preserving the interface and causal observation boundary of Section~\ref{sec:design}. To our knowledge, this is the first use of a validated program-level serving simulator as the evaluator for automated agent-serving policy search.

Each candidate is a Python class implementing one policy interface from Equation~\ref{eq:interface}. Candidates may access PCB state, the current time, and KV-pool utilization, but not future trace information. Thus, an evolved policy uses only information available in deployment and executes unchanged in vLLM. We instantiate the search loop using OpenEvolve~\citep{openevolve}, which manages the candidate population and LLM-generated code modifications, and run it for 50 iterations on each policy axis. Fitness is the ratio of the seed policy's mean JCT to the candidate's mean JCT on the target cell, so 1.0 indicates parity with the seed and higher values indicate better performance. The target cell serves Llama-3.1-70B at TP2 on RTX PRO 6000 with 30 SWE-bench programs arriving at 0.1 JPS. Each evaluation is one simulator run of the cell and takes a median of 19 minutes on one CPU core. Complete search settings and costs are reported in Appendix~\ref{app:evolved}.

\begin{wraptable}{r}{0.48\linewidth}
\vspace{-10pt}
\caption{Evolutionary policy search after 50 iterations per axis. Values are mean JCT for each hand-written seed and the best evolved policy; larger reduction is better.}
\label{tab:search}
\centering
\scriptsize
\setlength{\tabcolsep}{3pt}
\begin{tabular}{lrrr}
\toprule
axis & seed policy JCT & evolved policy JCT & reduction \\
\midrule
retention & 2534.3~s & 2522.7~s & $0.5$\% \\
scheduling & 2619.4~s & 2545.8~s & $2.8$\% \\
\bottomrule
\end{tabular}
\end{wraptable}

Table~\ref{tab:search} compares the best evolved policies with their hand-written seeds. On the retention axis, the search starts from TTL retention with $\tau=2$~s~\cite{li2026continuum}. The best evolved policy estimates the gap for each tool using an exponential moving average, scales the protection window with context length, and shortens it as KV-pool utilization increases:
\begin{equation}
d_{p,k}
=
c_{p,k}
+
s(n_{p,k})\,\hat g_{z_{p,k}}\,h(u),
\label{eq:evolved-retention}
\end{equation}
where $\hat g_z$ is the running gap estimate for tool $z$, $s(n)=\max(0.5,n/(2\times10^4))$ for context length $n$, and $h(u)=\max(0.3,1-u)$ for KV-pool utilization $u$ at call completion. The evolved policy reduces mean JCT by 0.5\% relative to the TTL seed.

On the scheduling axis, the search starts from PLAS~\cite{luo2025autellix}. The best evolved policy replaces attained service with its square root and incorporates context length and KV residency:
\begin{equation}
q_{p,k}
=
10^{3}\sqrt{A_p(r_{p,k})}
+
0.03\,n_{p,k}
-
0.10\,n_{p,k}\,
\mathbb{I}[\text{resident}].
\label{eq:evolved-sched}
\end{equation}
Lower $q_{p,k}$ runs first. The concave service term limits the penalty on programs that have already consumed substantial service, the context-length term favors calls requiring less prefill computation and KV capacity, and the residency term favors calls whose contexts remain cached. The evolved policy reduces mean JCT by 2.8\% on the target cell and by 2.3\% at 0.02 JPS. The hand-written seeds and evolved policies are provided in Appendix~\ref{app:evolved}. These policies are deployment-specific by design: the objective is to optimize serving decisions for a target workload and hardware configuration, rather than to learn one policy that generalizes across deployments.

\section{Conclusion}
\label{sec:conclusion}

We presented \sysname{}, a program-level simulator for agent serving in which successor arrivals and KV residency are outcomes of simulated execution rather than fixed trace inputs. A Program Orchestrator and Program Control Blocks maintain execution across calls, while the Retention and Dispatch Planes expose programmable KV-management, routing, and scheduling decisions through a common interface shared with vLLM. Across 20 paired simulator--real cells, mean-JCT error remains within 5.5\% on B200 and 5.2\% on RTX PRO 6000, including under memory pressure. Beyond validation, \sysname{} enables controlled exploration of interacting serving policies and serves as a CPU-based evaluator for LLM-driven evolutionary policy search, producing retention and scheduling policies that improve mean JCT over their hand-written seeds.

\bibliographystyle{plainnat}
\bibliography{references}
\newpage
\appendix
\section{Workloads}
\label{app:workloads}

Traces were collected by running mini-SWE-agent~\citep{minisweagent2025} on SWE-bench~\citep{jimenez2024swebench} instances and a multi-turn harness on BFCL~v4~\citep{patil2025bfcl}, driving an API-hosted model. Each turn records its prompt token count, output token count, tool identity, and wall-clock tool duration; nothing else is used. The served model in every experiment is Llama-3.1, and replay pins the recorded token counts, so the trace supplies request shapes and tool gaps, not text. Output lengths and tool durations are read by the replay harness only; policies see the PCB.

\begin{table}[h]
\caption{Trace statistics. Context tokens are per turn (each turn's prompt contains its predecessors').}
\centering
\small
\setlength{\tabcolsep}{4pt}\begin{tabular}{lrllll}
\toprule
& programs & turns/program & context tok/turn & output tok/turn & tool gap (s) \\
& & mean / med / max & mean / med / max & mean / med & mean / med / p99 \\
\midrule
SWE-bench & 50 & 57.5 / 47 / 185 & 13,463 / 11,115 / 52,724 & 94 / 52 & 1.09 / 0.20 / 18.6 \\
BFCL v4 & 100 & 7.0 / 5.5 / 21 & 6,603 / 3,448 / 71,370 & 42 / 25 & 1.39 / 1.06 / 5.7 \\
\bottomrule
\end{tabular}
\end{table}

Arrival times are Poisson at 0.02, 0.04, 0.06, 0.08, and 0.1 JPS, the same schedule on both sides. The SWE-bench trace totals 38.7M context tokens; at 0.02 JPS the arrival span is 2,004 s.

\section{Per-cell results}
\label{app:cells}

\begin{figure}[h]
\centering
\includegraphics[width=\linewidth]{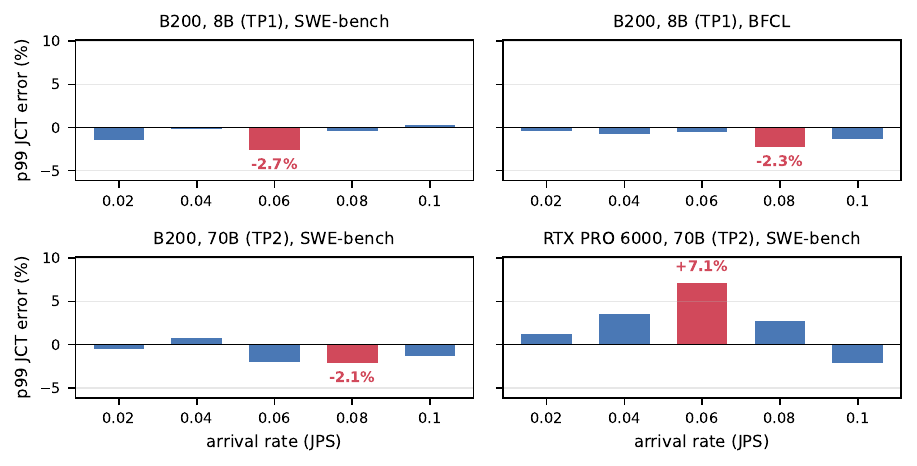}\\[6pt]
\includegraphics[width=\linewidth]{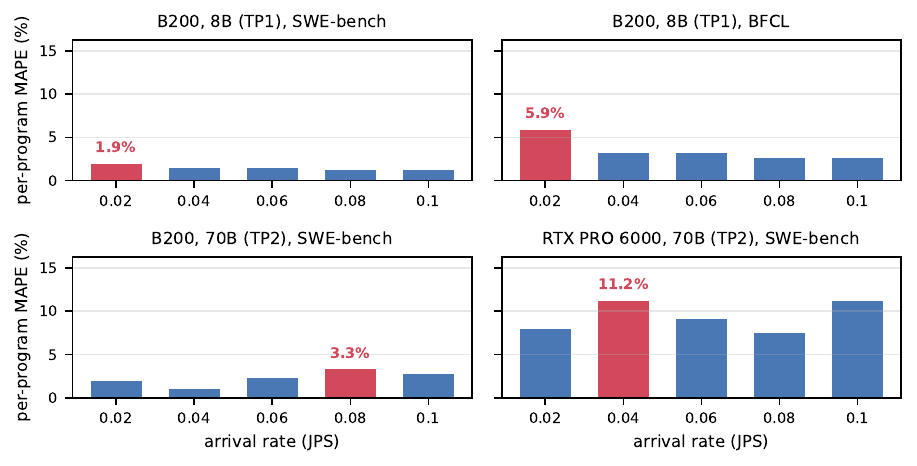}
\caption{p99-JCT error (top) and per-program MAPE (bottom) of \sysname{} relative to real vLLM under TTL retention and PLAS scheduling, per arrival rate; the largest value in each configuration is marked.}
\label{fig:appendix-err}
\end{figure}

Mean-JCT error, p99-JCT error, and per-program MAPE for every paired cell (sim minus real), all under TTL retention and PLAS scheduling; $\tau = 2$ s, the value Continuum measures from tool-call durations.

\begin{table}[h]
\centering
\scriptsize
\begin{tabular}{llrrrrrr}
\toprule
cell & rate & sim mean (s) & real mean (s) & mean err & p99 err & MAPE & p90 $|$err$|$ \\
\midrule
B200 8B SWE & 0.02 & 86 & 87 & $-1.4$ & $-1.5$ & 1.9 & 2.9 \\
 & 0.04 & 87 & 88 & $-1.3$ & $-0.1$ & 1.5 & 4.8 \\
 & 0.06 & 90 & 91 & $-1.6$ & $-2.7$ & 1.5 & 3.5 \\
 & 0.08 & 91 & 92 & $-0.8$ & $-0.4$ & 1.3 & 3.5 \\
 & 0.10 & 92 & 92 & $-0.5$ & $+0.3$ & 1.2 & 3.2 \\
B200 8B BFCL & 0.02 & 11 & 12 & $-5.5$ & $-0.4$ & 5.9 & 16.0 \\
 & 0.04 & 11 & 11 & $-2.7$ & $-0.8$ & 3.2 & 5.7 \\
 & 0.06 & 11 & 11 & $-2.6$ & $-0.5$ & 3.2 & 6.2 \\
 & 0.08 & 11 & 11 & $-2.7$ & $-2.3$ & 2.6 & 4.5 \\
 & 0.10 & 11 & 11 & $-2.5$ & $-1.3$ & 2.6 & 3.4 \\
B200 70B SWE & 0.02 & 153 & 155 & $-1.0$ & $-0.5$ & 1.9 & 4.7 \\
 & 0.04 & 159 & 160 & $-0.2$ & $+0.8$ & 1.0 & 2.6 \\
 & 0.06 & 166 & 170 & $-2.2$ & $-2.1$ & 2.3 & 4.6 \\
 & 0.08 & 173 & 179 & $-3.5$ & $-2.1$ & 3.3 & 8.6 \\
 & 0.10 & 178 & 182 & $-2.7$ & $-1.3$ & 2.8 & 5.5 \\
RTX 70B SWE & 0.02 & 3613 & 3495 & $+3.4$ & $+1.3$ & 8.0 & 11.8 \\
 & 0.04 & 4167 & 4096 & $+1.7$ & $+3.6$ & 11.2 & 24.3 \\
 & 0.06 & 5110 & 4888 & $+4.5$ & $+7.1$ & 9.1 & 16.8 \\
 & 0.08 & 5273 & 5101 & $+3.4$ & $+2.8$ & 7.4 & 15.1 \\
 & 0.10 & 5521 & 5825 & $-5.2$ & $-2.1$ & 11.2 & 25.7 \\
\bottomrule
\end{tabular}
\end{table}

\section{Request-level Simulation}
\label{app:ablation}

The program layer is inactive for single-turn requests and should cost nothing. Table~\ref{tab:sharegpt} compares \sysname{} with real vLLM~0.19 on ShareGPT (300 requests at 10 requests/s, B200, Llama-3.1-8B at TP1): mean latency matches to $+2.8$\% and p99 latency to $+2.0$\%, TPOT to $+2.8$\%, TTFT to $+5.9$\% (median $+1.5$\%), and generated-token throughput to $-0.1$\%.

\begin{table}[h]
\caption{Request-level workload: ShareGPT, 300 requests at 10 requests/s, B200, Llama-3.1-8B at TP1. Errors are sim relative to real.}
\label{tab:sharegpt}
\centering
\footnotesize
\setlength{\tabcolsep}{5pt}\begin{tabular}{lrrrrrrrrr}
\toprule
& \multicolumn{3}{c}{mean} & \multicolumn{3}{c}{p50} & \multicolumn{3}{c}{p99} \\
\cmidrule(lr){2-4}\cmidrule(lr){5-7}\cmidrule(lr){8-10}
metric & real & sim & err & real & sim & err & real & sim & err \\
\midrule
latency (s) & 3.17 & 3.26 & $+2.8$\% & 3.07 & 3.15 & $+2.8$\% & 4.03 & 4.10 & $+2.0$\% \\
TTFT (ms) & 17.96 & 19.01 & $+5.9$\% & 16.23 & 16.48 & $+1.5$\% & 51.96 & 52.17 & $+0.4$\% \\
TPOT (ms) & 4.84 & 4.97 & $+2.8$\% & 4.89 & 5.09 & $+3.9$\% & 5.21 & 5.33 & $+2.3$\% \\
\bottomrule
\end{tabular}
\end{table}

\section{TTFT, TPOT and Throughput Validation}
\label{app:percall}

\begin{figure}[h]
\centering
\begin{minipage}[t]{0.45\linewidth}
\centering
\includegraphics[width=\linewidth]{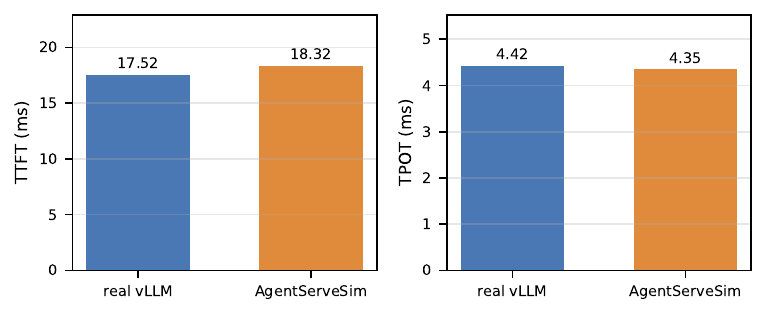}
\caption{B200 8B SWE-bench at 0.04 JPS under TTL retention and PLAS scheduling: mean TTFT (left) and TPOT (right) over 2,873 calls, real vLLM against \sysname{}.}
\label{fig:cell}
\end{minipage}\hfill
\begin{minipage}[t]{0.52\linewidth}
\centering
\includegraphics[width=\linewidth]{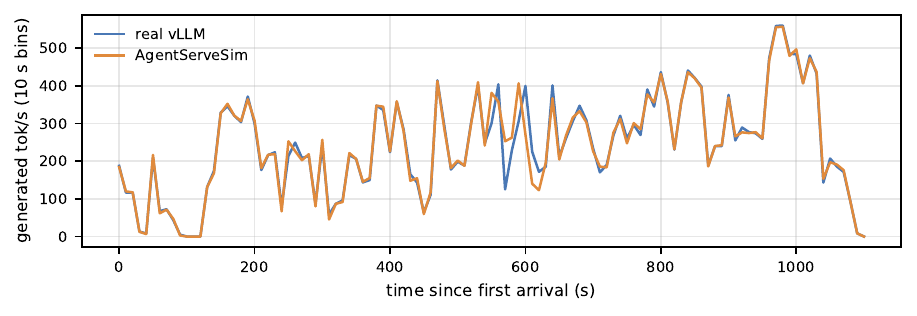}
\caption{Same cell: instantaneous generated-token throughput in 10 s bins, computed identically on both sides from per-call token counts and first- and last-token times.}
\label{fig:cell-tput}
\end{minipage}
\end{figure}

Figures~\ref{fig:cell} and \ref{fig:cell-tput} show per-call metrics for one cell, B200 8B SWE-bench at 0.04 JPS. Mean TTFT is 18.3 ms simulated against 17.5 ms real and mean TPOT 4.35 against 4.42 ms. The TTFT offset of about 0.8 ms is the mean wait for the in-flight decode step, which the simulator charges to every arriving call; it does not grow with load and contributes 0.3\% to JCT. Instantaneous throughput tracks the real trace bin by bin: correlation 0.98, mean absolute error 6\% per 10 s bin, and the same 1,100 s makespan. Generated-token throughput over the whole run is within 1.7\% of real on every validation cell.

\section{Policy interfaces}
\label{app:policies}

A retention policy implements \texttt{on\_turn\_complete(pcb, now)} and \texttt{on\_turn\_arrival(pcb, now)}, each returning a decision among \texttt{protect(until)}, \texttt{release}, \texttt{evict}, or \texttt{none}; a scheduling policy implements \texttt{priority(pcb, now)} returning an integer (smaller runs first, matching vLLM); a routing policy implements \texttt{select(pcb, inflight, now)} returning an instance index. The shipped retention policies are evict-always, cache-LRU, TTL ($\tau$ fixed), pressure-scaled TTL, Continuum per-tool TTL (0.9 quantile of observed gaps for the tool, default 2 s), and min-waste; scheduling: FCFS, program-FCFS, PLAS; routing: round-robin, least-loaded, session affinity.

The pressure-scaled TTL used in the experiments follows SAGA~\cite{guo2026saga}: with $u$ the KV-pool utilization at completion,
\begin{equation}
g(u) = 1 - 0.5\, m(u), \qquad m(u) = \min\!\left(1, \max\!\left(0, \frac{u - 0.7}{0.9 - 0.7}\right)\right),
\end{equation}
so the deadline shrinks linearly from $\tau$ at 70\% utilization to $\tau/2$ at 90\%. Min-waste is InferCept's retain-versus-discard test. Let $n_{p,k}$ be the context length in tokens, $\hat g_{p,k}$ the predicted gap (per-tool quantile, default 1 s), and $W_{\mathrm{discard}}(n_{p,k})$ InferCept's Equation~4: the token-seconds wasted by evicting now and re-prefilling at the next call in chunks of $c_h = \max(S - n_{\mathrm{inflight}}, 1)$, where $S$ is the chunked-prefill budget and $n_{\mathrm{inflight}}$ the tokens in flight, including the slowdown imposed on the running batch, with forward-pass time $t(n) = (a n + c)/1000$ s from a per-device profile $(a, c, S)$. The policy retains ($d_{p,k} = c_{p,k} + \hat g_{p,k}$) if $\hat g_{p,k}\, n_{p,k} \le W_{\mathrm{discard}}(n_{p,k})$ and evicts at completion otherwise; the shipped instance sets no value, so among retained entries reclamation falls back to the deadline order.

\section{Seed and evolved policies}
\label{app:evolved}

\paragraph{Search settings.} OpenEvolve runs 50 iterations with a population of 40 in two islands (migration every 10 iterations, elite ratio 0.2, exploration 0.3, exploitation 0.6), diff-based code mutations proposed by an ensemble of Qwen3.7-plus (weight 0.7) and Claude Sonnet~5 (weight 0.3) at temperature 1.0, and a 90-minute evaluation timeout; the prompt shows the three best and two diverse programs of the population. The retention search evaluated 48 candidates, of which 40 land within 0.5\% of one another, since on the target cell the deadline rarely decides which entry is reclaimed. The scheduling search scores each candidate on the target cell and then at 0.02 JPS; the context-length and residency terms of Equation~\ref{eq:evolved-sched} appeared in the first generation and account for most of the gain, and the square root arrived in the last iteration.

Listings~\ref{lst:seed-ret} to \ref{lst:evo-sched} give the policy bodies of Section~\ref{sec:usecase}: the region between the search markers of each candidate file, with docstrings removed. Both seeds and both evolved policies read only PCB fields, the current time, and the KV-utilization signal.

\begin{lstlisting}[caption={Seed retention policy: TTL, $\tau = 2$ s~\citep{li2026continuum}.},label={lst:seed-ret}]
class EvolvedRetention(RetentionPolicy):
    TAU_S = 2.0

    def on_turn_complete(self, pcb, request_id, now):
        return ("protect", now + self.TAU_S)

    def on_turn_arrival(self, pcb, now):
        return "release"
\end{lstlisting}

\begin{lstlisting}[caption={Evolved retention policy (Equation~\ref{eq:evolved-retention}).},label={lst:evo-ret}]
class EvolvedRetention(RetentionPolicy):
    TAU_S = 2.0

    def __init__(self):
        self.gap_ema = {}   # aggregate state: EMA of observed gaps per tool

    def on_turn_complete(self, pcb, request_id, now):
        est = self.gap_ema.get(pcb.tool_name, self.TAU_S)
        # larger contexts cost more to re-prefill: 10k tokens -> 0.5x, 20k -> 1x, 40k -> 2x
        scale = max(0.5, pcb.context_tokens / 20000.0)
        tau = est * scale
        u = self.signals.kv_utilization
        if u is not None:
            tau *= max(0.3, 1.0 - u)
        return ("protect", now + tau)

    def on_turn_arrival(self, pcb, now):
        g = pcb.gap_elapsed_s(now)
        if g is not None and pcb.tool_name is not None:
            p = self.gap_ema.get(pcb.tool_name)
            self.gap_ema[pcb.tool_name] = g if p is None else 0.8 * p + 0.2 * g
        return "release"
\end{lstlisting}

\begin{lstlisting}[caption={Seed scheduling policy: PLAS~\citep{luo2025autellix}.},label={lst:seed-sched}]
class EvolvedScheduling(SchedulingPolicy):
    engine_args = {"scheduling_policy": "priority"}

    def priority(self, pcb, now):
        return int(round(pcb.attained_service_s * 1000.0))
\end{lstlisting}

\begin{lstlisting}[caption={Evolved scheduling policy (Equation~\ref{eq:evolved-sched}).},label={lst:evo-sched}]
class EvolvedScheduling(SchedulingPolicy):
    engine_args = {"scheduling_policy": "priority"}

    PREFILL_MS_PER_TOKEN = 0.03    # proxy for prefill hold time on the KV pool
    RESIDENT_MS_PER_TOKEN = 0.10   # value of skipping re-prefill when KV is resident

    def priority(self, pcb, now):
        # concave PLAS: sqrt(attained service) bounds how far a long program falls behind
        base = math.sqrt(max(0.0, pcb.attained_service_s)) * 1000.0
        # smaller context -> admitted first
        size_term = pcb.context_tokens * self.PREFILL_MS_PER_TOKEN
        # resident context -> admitted first; scales with the avoided re-prefill
        kv_bonus = (pcb.context_tokens * self.RESIDENT_MS_PER_TOKEN
                    if pcb.kv_protected else 0.0)
        return int(round(base + size_term - kv_bonus))
\end{lstlisting}

\newpage

\end{document}